# Theoretical Foundations for Abstraction-Based Probabilistic Planning

Vu Ha    Peter Haddawy*
Department of EE & CS
University of Wisconsin-Milwaukee
{vu, haddawy}@cs.uwm.edu

## Abstract

Modeling worlds and actions under uncertainty is one of the central problems in the framework of decision-theoretic planning. The representation must be general enough to capture real-world problems but at the same time it must provide a basis upon which theoretical results can be derived. The central notion in the framework we propose here is that of the *affine-operator*, which serves as a tool for constructing (convex) sets of probability distributions, and which can be considered as a generalization of belief functions and interval mass assignments. Uncertainty in the state of the worlds is modeled with sets of probability distributions, represented by *affine-trees*, while actions are defined as tree-manipulators. A small set of key properties of the affine-operator is presented, forming the basis for most existing operator-based definitions of probabilistic action projection and action abstraction. We derive and prove correct three projection rules, which vividly illustrate the precision-complexity tradeoff in plan projection. Finally, we show how the three types of action abstraction identified by Haddawy and Doan are manifested in the present framework.

## 1 INTRODUCTION

In the framework of decision-theoretic planning (DTP), uncertainty in the state of the world and in the effects of actions are represented with probabilities, and the planner's goals, as well as tradeoffs among them, are represented with utilities. Given this representation, the objective is to find an optimal or near optimal plan or policy, where optimality is defined as maximizing expected utility.

In most of the existing DTP approaches, the world is represented with a probability distribution over the state space, and actions are defined as stochastic mappings among the states [12, 3, 1, 16]. Given this framing of the problem, all probabilistic and decision-theoretic planners face the burden of computational complexity in seeking an optimal or near-optimal solution. One popular way to address this problem is to use abstraction techniques to guide the search through the plan space and to reduce the cost of plan evaluation. This concept has been applied in Markov process-based planning [1] as well as less structured approaches [12, 9]. Because of the wide applicability of abstraction techniques in decision-theoretic planning, a simple and general theory of action abstraction that could form the basis for comparison of approaches and development of new approaches is desirable. This paper attempts to provide such a theory.

This paper has its origins in the work of Hanks [12], Tenenberg [14], and Haddawy and Suwandi [11], as well as later work by Doan and Haddawy [8, 9, 4, 5]. In the framework of Doan and Haddawy, the planning problem is formalized as the problem of searching for the optimal plan through the space of possible plans, each of which is a sequence of actions. The planner is armed with the capability of understanding and evaluating abstract plans constructed from abstract actions, and thus is able to evaluate and eliminate a set of suboptimal plans without explicitly evaluating each member of that set. The introduction of abstract actions, however, discards the conventional world and action representations. An abstract action is a function that maps a probability distribution to a *set* of probability distributions.

A mechanism for representing sets of probability functions in this planning framework must satisfy the following requirements. First, it must be computationally achievable; we cannot, for example, just explicitly list all the members of the set. Second, it must enable the planner to project actions, especially abstract actions on the worlds it represents. Third, since pro-

*Thanks to AnHai Doan for helpful comments. This work was partially supported by NSF grant IRI-9509165.

jecting an abstract plan will result in a set of probability functions, transforming the task of computing expected utilities to the task of computing expected utility *intervals*, algorithms to perform this task must be provided. And finally, the mechanism should support the process of abstracting actions and plans.

One of the first attempts at addressing the first two of the above four problems is the work of Chrisman [2]. Chrisman represents uncertainty by a set of probability functions that are consistent with a belief function, which in turn is represented by a basic probability function (or a mass assignment). He then provides a closed-form projection rule to project actions on this kind of set. Doan [4, 5] attempts to answer all of the four questions posted in the preceding paragraph by introducing the notion of *general* or *interval mass assignment* (IMA), which is used as a representation for certain (convex) sets of probability functions. Although introduced as a generalization of belief functions, IMAs are still not expressive enough; in order to derive a closed-form projection rule, we must sacrifice too much information and thus obtain weak approximations for the projected worlds.

In this paper, we address the above problems by introducing the *affine-operator*, which is used to construct (convex) sets of probability functions, including (Theorem 1) IMAs and belief functions as special cases. The worlds are represented as *affine-trees*, which dynamically grow with the advancing of the projection process. Actions are categorized as *primitive* - mapping probability functions to probability functions, and *abstract* - mapping probability functions to sets of probability functions.

The rest of this paper is organized as follows. In Section 2.1, we define the affine-operator, affine-trees, and the class of affine-worlds. The heart of our framework is a small set of lemmas investigating the properties of affine-trees and is presented in Section 2.2. We present the action model in Section 3. Three projection rules are presented and shown to be correct in Section 4 (Theorem 2). The projection rules are descending in term of precision but ascending in term of practical usefulness. The problem of computing the expected utility interval is solved by an efficient algorithm suggested by Theorem 3 in Section 5. In Section 6, we revisit the three abstraction techniques identified by Haddawy and Doan and prove them correct (Theorem 4). We then conclude and discuss future research in Section 7. Due to space limitations, proofs of several theorems are only sketched or omitted. The complete proofs, as well as additional discussion can be found in [7].

**Terminology and Notation.** We will use the following notations throughout the paper. A *state* $s$ represents a possible complete description of all information relevant to the planner. The set of all such descriptions is call the *state space* (denoted $\Omega$) and is assumed to be finite. A *world* $w$ is the uncertain knowledge the planner actually has at a certain time point, which can be either (i) a state $s \in \Omega$ (no uncertainty, deterministic world), (ii) a Bayesian [1] probability distribution over the state space $\Omega$ (probabilistic world), or (iii) a set of such probability distributions (general world). The set of all probabilistic worlds is denoted by $\wp$, and thus the set of all general worlds can be denoted by $2^\wp$. The above three classes of worlds form a chain in which each element is a strict superset of its preceding element: $\Omega \subset \wp \subset 2^\wp$. For the sake of simplicity, the interpretation of a world (a state, a PD, or a set of PDs) will be left implicit in the discussion where misunderstanding can be excluded. A world $w_1 \in 2^\wp$ is said to *subsume* a world $w_2 \in 2^\wp$ if $w_1 \supseteq w_2$. The *convex hull* of a world $w \in 2^\wp$, denoted by $\mathbf{CH}(w)$ is defined to be the smallest world that contains $w$ and all convex combinations [2] of elements of $w$. A world $w$ is said to be *convex*, if $w = \mathbf{CH}(w)$.

We will be performing operations on intervals. Every interval is assumed to be an (open or closed) subinterval of the interval $[0, 1]$. The set of all such intervals is denoted by $\mathcal{Q}$. Operations on the intervals such as addition, multiplication, etc. are defined in the obvious way. For example, the multiplication of two intervals $[l_1, u_1]$ and $[l_2, u_2]$ is defined to be the interval $[l_1.l_2, u_1.u_2]$. The lower and upper bound of an interval $Q$ are denoted by $L(Q)$ and $U(Q)$, respectively.

## 2 THE AFFINE-OPERATOR AND AFFINE-WORLDS

### 2.1 BASIC DEFINITIONS

We first introduce the notion of *affine-vectors*. Any vector whose components are numbers between 0 and 1 and sum up to 1 is called an *affine-vector* [3]. The central notion of our framework is the notion of the *affine-operator*, defined as follows.

**Definition 1 (The Affine-Operator)** *The affine operator defined by an n-dimension affine-vector $\vec{q} = (q_1, q_2, \ldots, q_n)$ is the function that maps any vector $\vec{P} = (P_1, P_2, \ldots, P_n) \in \wp^n$ of probability functions to the probability function $R = \vec{q} \otimes \vec{P} := \sum_{i=1}^{n} q_i P_i$.*

It is not hard to see that the above definition is correct, i.e., $R$ is indeed a probability function. We extend the affine operator, "$\otimes$" to deal with vectors of intervals and vectors of worlds as follows.

---
[1] Bayesian probability distributions are probability functions that assign a probability number to every single element of the sample space (which in this case is $\Omega$).
[2] A convex combination of $n$ probability functions $P_1, P_2, \ldots, P_n$ is a probability function of the form $\sum_{i=1}^{n} \alpha_i P_i$, where $0 \leq \alpha_i \leq 1$ and $\sum_{i=1}^{n} \alpha_i = 1$.
[3] The "affine" term comes from the similar terminology used in linear algebra.

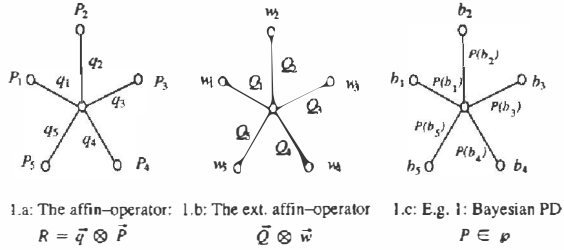

1.a: The affin-operator:  1.b: The ext. affin-operator  1.c: E.g. 1: Bayesian PD
$R = \vec{q} \otimes \vec{P}$                     $\vec{Q} \otimes \vec{w}$                               $P \in \wp$

Example 2: When all intervals are [0,1]:

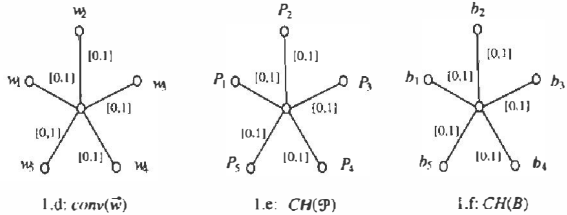

1.d: $conv(\vec{w})$        1.e: $CH(\mathcal{P})$        1.f: $CH(B)$

Figure 1: The affine-operator and examples

**Definition 2 (The Extended Affine-Operator)**
*The affine-operator defined by an n-dimension interval vector $\vec{Q} = (Q_1, Q_2, \ldots, Q_n) \in \mathcal{Q}^n$ is the function that maps any world vector $\vec{w} = (w_1, w_2, \ldots, w_n) \in (2^\wp)^n$ to the world $\vec{Q} \otimes \vec{w}$ that is the set of all probability functions of the form $\vec{q} \otimes \vec{P}$, where $\vec{q} = (q_1, q_2, \ldots, q_n)$ is any affine-vector such that $q_i \in Q_i$, and $\vec{P} = (P_1, P_2, \ldots, P_n)$ is any vector of probability functions such that $P_i \in w_i$, for all $i = 1, 2, \ldots, n$.*

Henceforth, the term affine-operator will be used to refer to the extended definition. The probability function $R = \vec{q} \otimes \vec{P}$ and the world $\vec{Q} \otimes \vec{w}$ can be represented by a center of a star $S$ with $n$ branches (and thus $n$ leaves) (Figures 1.a and 1.b). The branches are associated with the numbers $q_i$ and intervals $Q_i$, while the leaves are associated with probability functions $P_i$ and the worlds $w_i$, $i = 1, 2, \ldots, n$, respectively. We will call a star whose branches are associated with intervals an *affine-star*. Thus, every affine-star will define exactly one affine-operator.

**Example 1.** We first notice that any Bayesian probability function $P$ can be represented by an $|\Omega|$-branch affine-star (Figure 1.c). The leaves of this star are associated with the states $b \in \Omega$, and the branches are associated with the probability numbers $P(b)$.

**Example 2.** Consider the case where each branch $Q_i$ is the entire interval $[0,1]$ (Figures 1.d, 1.e, 1.f). The corresponding affine-operator will create the *convex combination* of the leaf-worlds $w_1, w_2, \ldots, w_n$, denoted by $conv(w_1, w_2, \ldots, w_n)$, or $conv(\vec{w})$ (Figure 1.d). When each world $w_i$ is a single probability function $P_i$, the world $conv(P_1, P_2, \ldots, P_n)$ is exactly the conventional convex hull of the set of probability functions $\mathcal{P} = \{P_1, P_2, \ldots, P_n\}$, $CH(\mathcal{P})$ (Figure 1.e). In the more special case, where each probability function $P_i$ is a state $b_i$, with $B = \{b_1, b_2, \ldots, b_n\}$, the convex hull of $B$, $CH(B)$ (Figure 1.f) is the set of all probability functions that assign positive probability to only the elements of $B$. Notice that the function that maps any set of states $B$ to the world $CH(B)$ is monoton increasing, i.e., $CH(B) \subset CH(C)$ iff $B \subset C$. In particular, $CH(\Omega) = \wp$. Furthermore, $conv(CH(B), CH(C)) = CH(B \cup C)$.

Sometimes it can be the case that in an affine-star some of the leaf-worlds $w_i$ are also worlds constructed by using the affine operator, and are represented by other affine-stars. This observation gives rise to the notion of affine-trees. A tree whose branches are associated with intervals is called an *affine-tree*. Affine-stars are thus special affine-trees with depth 1.

When the leaves of an affine tree $T$ are associated with worlds, then each node $N$ of $T$ will be associated with the world obtained by applying the affine operator recursively on the subtree of $T$ that has $N$ as its root. An affine-tree thus defines the composition of a sequence of affine-operators (or affine-stars).

Within the class of worlds that can be represented by affine-trees, we define the following class of worlds.

**Definition 3 (Affine-Worlds)** *The class of affine-worlds, denoted by $\mathcal{AFF}$, is the set of all worlds that can be represented by an affine-tree whose leaves are associated with states.*

If $w$ is an affine-world represented by an affine-tree $T$ whose leaves are associated with states, then $T$ is called a *standard affine-tree* of $w$. From Examples 1 (Figure 1.c), we know that any probability function is an affine-world having a standard affine-tree with depth 1. Example 2 (Figure 1.f) tells us that for any $B \subseteq \Omega$, $CH(B)$ also has the same properties. The following theorem says that belief functions [13], when interpreted as sets of probability functions, are also affine-worlds having standard affine-trees of depth 2.

**Theorem 1 (Belief Functions Are Affine-Worlds)**
*Let Bel be a belief function with the corresponding basic probability function (mass assignment) m. Denote the focal element of m by $B_1, B_2, \ldots, B_n$. Then the set of probability functions that are consistent [4] with Bel is an affine-world represented by an n-branch affine-star with branches $m(B_i)$ and leaves $CH(B_i)$.*

In [4, 5], belief functions, by means of mass assignments, are generalized to interval mass assignments by allowing mass functions to be interval-valued. It is clear that interval mass assignments are affine-worlds represented by affine-stars with interval-branches and leaves of the form $CH(B)$, $B \subseteq \Omega$.

---

[4] A probability function $P$ is consistent with a belief function $Bel$ if $Bel(B) \leq P(B), \forall B \subseteq \Omega$.

## 2.2 LEMMAS

In this section, we present a set of lemmas concerning the affine-operator that form the heart of our framework. In the rest of the paper, the leaves of any affine-tree will be assigned worlds. In order to make the discussion less cumbersome, we shall not always distinguish a branch of an affine-tree from the interval associated with it, or an affine-tree from the world it represents, when doing so does not introduce ambiguity. For example, the sentence "affine-tree $T_1$ subsumes affine-tree $T_2$" should be interpreted as meaning that the world represented by $T_1$ subsumes the world represented by $T_2$.

**Lemma 1 (Monotonicity Lemma)** *Replacing any leaf-world with a subsuming world, or any branch-interval with a subsuming interval in an affine-tree will result in a subsuming affine-tree.*

**Lemma 2 (Star-Merging Lemma)** *Consider a set of affine-stars $\{S_j | j \in J\}$ having the same number $n$ of branches. If $S$ is an $n$-branch affine-star whose branch-interval (respectively leaf-world) number $i$ subsumes branch-interval (respectively leaf-world) number $i$ of each $S_j, \forall j \in J, \forall i = 1, 2, \ldots, n$, then $S \supseteq S_j, \forall j \in J$.*

**Lemma 3 (Branch-Merging Lemma)** *Let $S$ be an affine-star with $n$ branches $Q_i$ and $n$ leaves $w_i$, and $J \subseteq \{1, 2, \ldots, n\}$. If $S'$ is the affine-star obtained from $S$ by "merging" branches index $j \in J$ to a single branch $\sum_{j \in J} Q_j$ with the corresponding leaf $conv(w_j | j \in J)$, then $S' \supseteq S$.*

**Lemma 4 (Tree-Flatening Lemma)** *Let $T$ be an affine-tree. If $T'$ is the tree obtained by "flatening" $T$, i.e., by replacing every path from the root $r$ to a leaf $l$ with a single branch associated with the multiplication of the branches along this path, then $T' \supseteq T$.*

**Lemma 5 (CH Invariance Lemma)** *Let $S$ be an affine-star with $n$ branches $Q_i$ and $n$ leaves $w_i$, i.e. $S = \vec{Q} \otimes \vec{w}$. Then $\mathbf{CH}(S) = \vec{Q} \otimes \mathbf{CH}(\vec{w})$ [5], where $\mathbf{CH}(\vec{w}) = (\mathbf{CH}(w_1), \mathbf{CH}(w_2), \ldots, \mathbf{CH}(w_n))$.*

**Corollary 1** $\mathbf{CH}(w) = w, \forall w \in \mathcal{AFF}$. *As a consequence, affine-worlds are convex sets of probability functions.*

## 3 ACTION MODEL

### 3.1 PRIMITVE ACTIONS

A *primitive action* $\lambda$ is a function mapping states into probability distributions over states: $\lambda : \Omega \to \wp$, and is represented by a finite set of tuples $\{<C_i, p_i, e_i>| i\}$ called *branches*. In each branch $i$, $C_i$ is a set of states, called the *condition*, $p_i$ is a number in $[0, 1]$, called

---

[5]Thanks to Vu Ha Van for proving the $\supseteq$ direction.

the *probability*, and $e_i$ is a function mapping states into states, called the *primitive effect* of action $\lambda$. The conditions $C_i$ must be jointly exhaustive, i.e. their union is $\Omega$. The semantics of action $\lambda$ is that it maps a state $b \in \Omega$ into a probability function $\lambda(b) := P_b \in \wp$ defined as: $P_b(a) = \sum_{i: b \in C_i; e_i(b) = a} p_i, \forall a \in \Omega$.

This semantics definition means that if the world before the execution of action $\lambda$ is $b$, then for each branch index $i$ whose condition satisfies $b$, the world after the execution of $\lambda$ will be $e_i(b)$ with probability $p_i$. It is clear that this semantic definition is correct only if for all $b \in \Omega$, the probabilities $P_b(a)$ for all $a \in \Omega$ sum to 1. By introducing the notion $S_b(C_i) = 1$ if $C_i \ni b$ and $S_b(C_i) = 0$ otherwise, this condition is equivalent to

$$\forall b \in \Omega : \sum_i S_b(C_i).p_i = 1. \qquad (1)$$

The probability function $\lambda(b) = P_b$ can then be represented by an $n$-branch affine-star whose branches are $S_b(C_i).p_i$ and whose leaves are $e_i(b)$, $i = 1, 2, \ldots, n$ (see Figure 2, the dash-lined ellipse). We extend action $\lambda$ (which is a $\Omega \to \wp$ function), to a $\wp \to \wp$ function as follows:

$$\forall P \in \wp : \lambda(P) := \sum_{b \in \Omega} P(b).\lambda(b).$$

It is not hard to see that $\lambda(P)$ is indeed a probability function represented by a 2-level affine-tree (Figure 2). In some similar frameworks, a probabilistic action is represented by a set of mutually exclusive and jointly exhaustive conditions [12, 11] or discriminants [1]. Each condition is then associated with a finite set of probability-effect pairs, where the sum of the probabilities is 1. An action in our framework can easily be transformed to this form [6]. Here we adopt this representation because it facilitates a natural generalization of primitive actions into abstract actions, as soon shown.

### 3.2 ABSTRACT ACTIONS

An abstract action $\Lambda$ is a function mapping worlds into worlds: $\Lambda : 2^\wp \to 2^\wp$, and is represented by a finite set of tuples $\{< C_i, P_i, E_i >| i\}$, where the $C_i$ are jointly exhaustive conditions, the $P_i$ are subintervals of $[0, 1]$, and the $E_i$ are functions mapping states into sets of states and are called *abstract effects*. In order to define the semantics of abstract actions, we first introduce the notion of effect and action instantiation.

---

[6]Define a binary relation $\sim$ on the state space $\Omega$ such that $\forall a, b \in \Omega : a \sim b$ iff $a$ and $b$ are satisfied by exactly the same set of conditions $C_i$. Clearly, $\sim$ is an equivalence relation on $\Omega$. The partition of $\Omega$ that corresponds to the factorization of $\Omega$ according to $\sim$ will then give us a collection of mutually exclusive and jointly exhaustive conditions. We can then reorganize the branches of the action according to these new conditions, obtaining the desired representation.

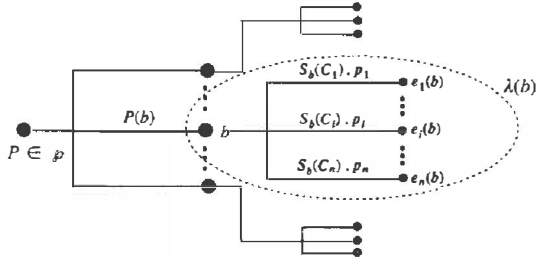

Figure 2: Semantics of primitive actions

A primitive effect $e$ is called an *instantiated effect* of an abstract effect $E$ (denoted $e \in E$) if for all $s \in \Omega$, we have $e(s) \in E(s)$. A primitive action $\lambda = \{<C_i, p_i, e_i>|i\}$ is called an *instantiated action* of an abstract action $\Lambda = \{<C_i, P_i, E_i>|i\}$ (denoted $\lambda \in \Lambda$) if $p_i \in P_i$ and $e_i \in E_i$ for each branch $i$. Note that the conditions $C_i$ of the instantiated action are the same as those of the abstract action. Primitive action $\lambda$, of course, must satisfy condition ( 1).

We are now in position to define the semantics of abstract actions. An action $\Lambda$ is a function that maps any world $w \in 2^\wp$ into the world

$$\Lambda(w) = \bigcup_{\lambda \in \Lambda; P \in w} \lambda(P).$$

Abstracting actions has long been a popular method to cope with the complexity of planning in large problem spaces. The models for abstract actions, however, are diverse, and we have not seen any work that models abstract actions as functions operating on sets of probability functions. In the MDP framework of Boutilier and Dearden [1], concrete states are clustered into abstract states according to a set of relevant attributes, and abstract actions are stochastic mappings among abstract states. In this sense, abstract actions are primitive actions wrt the abstract state space. In the work of Doan [4, 5], abstract actions do not have explicit semantics but are associated with projection rules. Defining actions as suggested in this paper offers some advantages in deriving procedures for abstracting actions, as we shall see later.

## 4    ACTIONS ON AFFINE-WORLDS

**Lemma 6 (Action Semi-Invariance Lemma)**
Let $\vec{Q} \in \mathcal{Q}^n$ be an n-dimension vector of intervals and $\vec{w} \in (2^\wp)^n$ be an n-dimension vector of worlds and $\Lambda$ be an action. Let $\Lambda(\vec{w}) = (\Lambda(w_1), \Lambda(w_2), \ldots, \Lambda(w_n))$. Then $\Lambda(\vec{Q} \otimes \vec{w}) \subseteq \vec{Q} \otimes \Lambda(\vec{w})$.

The Action Semi-Invariance Lemma validates a classical use of the "divide and conquer" technique in projecting actions, provided that the pre-action world can be represented by an affine-tree. To be more specific, let us consider a pre-action world represented by an affine-tree $T$. If we replace every leaf-world $l$ of the tree $T$ with the world $\Lambda(l)$, then the resulting affine-tree will represent a world that subsumes $\Lambda(w)$.

In the rest of the discussion, we shall assume that $\Lambda$ is an action that has $n$ branches: $\Lambda = \{(C_i, P_i, E_i)|i = 1, 2, \ldots, n\}$, and $w \in \mathcal{AFF}$ is an affine-world represented by an affine-tree $T(w)$. When $T(w)$ is in the standard form, i.e., the leaves of $T(w)$ are associated with states, then we can project action $\Lambda$ on $w$ by replacing every leaf-state $b$ with the world $\Lambda(b)$. This is exactly the way the first projection rule works.

**Projection Rule 1 (PR1)** *For every leaf $b \in \Omega$ of the affine-tree $T(w)$, we create an affine-star $T_1(\Lambda, b)$ (Figure 3, left) having $n$ branches (leaves). The branches are associated with the intervals $S_b(C_i).P_i$, while the leaves are associated with the worlds $E_i(b)$. The projected world is represented by the affine-tree $T_1(\Lambda, w)$, which is obtained from $T(w)$ by replacing every leaf $b$ with the corresponding affine-star $T_1(\Lambda, b)$.*

It is clear that if each leaf-world $E_i(b)$ of the affine-stars $T_1(\Lambda, b)$ is a singleton state (for example, if the $E_i$ are primitive effects), then the projected world $T_1(\Lambda, w)$ is also an affine-world. This will no longer be true if some effect $E_i$ maps a state $b$ into a set of at least two states. In other words, PR1, in general, is not closed on the class of affine-worlds, $\mathcal{AFF}$. In order to obtain a closed-form projection rule, we have to trade some precision. Notice that the best affine-approximation of a world $B \subseteq \Omega$ is its convex hull, **CH**$(B)$. This observation leads us to the second projection rule.

**Projection Rule 2 (PR2)** *Same as PR1, except that $T_1(\Lambda, b)$ is replaced by $T_2(\Lambda, b)$ (Figure 3, right) whose leaves are associated with **CH**$(E_i(b))$ instead of $E_i(b)$. The resulting affine-tree is denoted by $T_2(\Lambda, w)$.*

Note that according to the **CH** Invariance Lemma, we have that $T_2(\Lambda, w)$ is exactly the convex hull of $T_1(\Lambda, w)$, from which the correctness of PR2 is implied. Compared to PR1, PR2 gives a looser but more "representable" result. By forcing the projected worlds to always be affine-worlds, the projection process can continue with more and more actions, "growing" a projection tree with more and more levels and leaves. This corresponds to the well-known forward projection algorithm for probabilistic actions [12]. The next question then arises: What is the complexity of this process? The complexity of an affine-tree $T$ is estimated (in this discussion) by two factors: the number of the levels (or the depth) of $T$, denoted by $\mathcal{D}(T)$, and the number of the leaves of $T$, denoted by $\mathcal{L}(T)$.

Recall that in PR2, we replace every leaf $b$ with an affine-star having $n$ leaves **CH**$(E_i(b))$, $i = 1, 2, \ldots, n$ (see Figure 3, right). In order to apply PR2 to another action, we have to standardize $T_2(\Lambda, w)$, which amounts to standardizing **CH**$(E_i(b))$, for every leaf $b$ of $T(w)$ and $i = 1, 2, \ldots, n$. It is then clear that after each action projection, the standard affine-tree representing the current world will have two more levels: $\mathcal{D}(T_2(\Lambda, w)) = \mathcal{D}(T(w)) + 2$. Furthermore, if, on average, an abstract effect function $E_i$ maps a state into a set of $k$ states, then, on average, the number of the leaves of the current tree will increase by the factor of $n \times k$: $\mathcal{L}(T_2(\Lambda, w)) = n \times k \times \mathcal{L}(T(w))$.

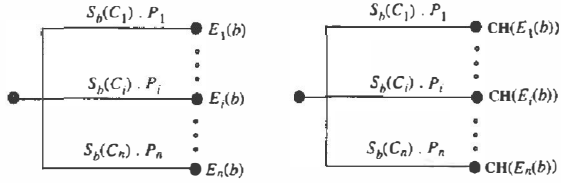

Figure 3: $T_1(\Lambda, b)$ (left), and $T_2(\Lambda, b)$ (right)

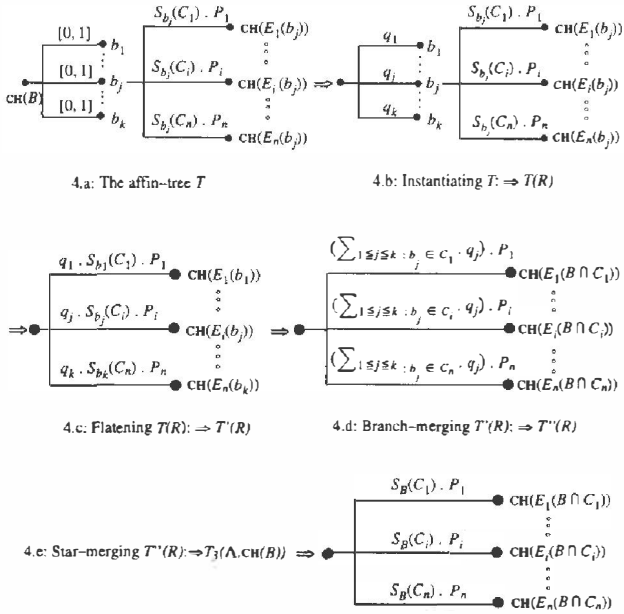

Figure 4: Illustration of the correctness proof of PR3

In order to make the projection process less complex, we have to trade some more precision. The third projection rule is devised with this goal in mind. In the third projection rule, we do not require that any affine-tree be standardized: it is sufficient to make sure that the leaves of the affine-trees be associated with worlds of the form $\mathbf{CH}(B)$, $B \subseteq \Omega$. We first introduce the following notation. For $B, C \subseteq \Omega$, let $S_B(C) = 1$ if $C \supseteq B$, 0 if $C \cap B = \emptyset$, and $[0, 1]$ otherwise.

**Projection Rule 3 (PR3)** *For each leaf $\mathbf{CH}(B)$ of the affine-world $T(w)$, we create an affine-star $T_3(\Lambda, \mathbf{CH}(B))$ (Figure 4.e) that has n branches (leaves). The branches are associated with the intervals $S_B(C_i).P_i$, while the leaves are associated with the worlds $\mathbf{CH}(E_i(B \cap C_i))$, $i = 1, 2, \ldots, n$. The affine-tree $T_3(\Lambda, w)$ is obtained from $T(w)$ by replacing every leaf $\mathbf{CH}(B)$ with the corresponding affine-star $T_3(\Lambda, \mathbf{CH}(B))$.*

It is clear that $\mathcal{D}(T_3(\Lambda, w)) = \mathcal{D}(T(w)) + 1$ and $\mathcal{L}(T_3(\Lambda, w)) = n \times \mathcal{L}(T(w))$. Thus, in a projection process using PR3, after each action projection the number of the levels of the current affine-tree will increase by one, and the number of its leaves will increase by the factor of $n$, the number of the branches of the currently projected action. Furthermore, unlike the first two projection rules, PR3 essentially operates on *sets* of states instead of states, and thus is more tractable in cases of state spaces with large numbers of states. The following theorem proves the correctness of PR3.

**Theorem 2 (The Correctness Theorem)**
*For any action $\Lambda$ and affine-world $w$, $T_3(\Lambda, w)$ is also an affine-world and $T_3(\Lambda, w) \supseteq T_2(\Lambda, w)$ (and thus $= \mathbf{CH}(T_1(\Lambda, w)) \supseteq \mathbf{CH}(\Lambda(w)) \supseteq \Lambda(w)$).*

*Proof:*(Sketch) The first part of the theorem is trivial. To prove the second part, it is sufficient to show that $T_3(\Lambda, \mathbf{CH}(B)) \supseteq T_2(\Lambda, \mathbf{CH}(B))$, for any $B = \{b_1, b_2, \ldots, b_k\} \subseteq \Omega$. The main idea of the proof is the following. We start with the affine-tree $T_2(\Lambda, \mathbf{CH}(B))$ (Figure 4.a) and perform various operations on it (Figures 4.b-e). After each operation (except the first one), the new affine-tree will subsume the old one, according to the lemmas in Section 2.2.

The first operation is instantiation: we instantiate $T_2(\Lambda, \mathbf{CH}(B))$ by taking a $T_2(\Lambda, R)$ affine-tree, where $R \in \mathbf{CH}(B)$. Denote this tree by $T(R)$ (Figure 4.b). We then tree-flaten $T(R)$, obtaining $T'(R)$ (Figure 4.c), branch-merge $T'(R)$ (according to each index $i$), obtaining $T''(R)$ (Figure 4.d). Finally, we star-merge $T''(R)$ for all $R \in \mathbf{CH}(B)$, and use the observation that $\sum_{1 \leq j \leq k: b_j \in C_i} R(b_j) \in S_B(C_i), \forall R \in \mathbf{CH}(B)$ to obtain the final affine-tree $T_3(\Lambda, \mathbf{CH}(B))$ (Figure 4.e), completing the proof (by the Monotonicity Lemma). □

Chrisman [2] provides a closed-form projection rule that works for probabilistic actions and worlds that can be represented by belief functions (or mass assignments). The action representation is a special case of our representation, and thus his result is also a special case of our result, namely PR3. Doan [5] uses IMAs to represent worlds and gives a projection rule that differs from PR3 in that the projected affine-tree is always flatened to an affine-star. A comparison of these two rules is discussed in Section 7.

## 5 Computing the EUI of Affine-Worlds

Computing the expected utility of a plan is one of the most frequently performed oparations of a DTP planner. The upcoming theorem shows that computing the expected utility of affine-worlds can be done elegantly, again using affine-trees as decompisition tools.

We first define the notion of utility function. A function $f : \Omega \rightarrow R$ is called a *utility function*. The *expected utility* of a world $w \in 2^\wp$ is the set of real numbers defined as $EU(w) = \{\sum_{s \in \Omega} P(s).f(s) | P \in w\}$. The *expected utility interval* of $w$, denoted by $EUI(w)$ is defined to be the convex hull [7] of $EU(w)$.

**Theorem 3 (Affine-Worlds Utility Theorem)**
For all affine-world $w \in \mathcal{AFF}$, $EU(w) = EUI(w)$.

*Proof:*(Sketch) We prove this theorem using structural induction. Let $w$ be an affine-world represented by a standard affine-tree $T$. Clearly, for any leave-state $s$ of $T$, we have $EUI(s) = EU(s)$. Suppose now that for every node $C$ ($C = child$) on the $r$th level of $T$, we have computed $EUI(C)$ and $EU(C) = EUI(C)$. Then computing the EUI of any node $P$ ($P = parent$), (i.e., the lower and the upper bounds of the interval $EUI(P)$) on the $(r-1)st$ level is a special instance of the knapsack problem, which can be efficiently solved using the greedy algorithm [8]. It is also not hard to see that since $EU(C) = EUI(C)$ for every child $C$, we will get $EU(P) = EUI(P)$ for parent $P$. □

## 6  Abstracting Actions

In exchange for some loss of information, the Third Projection Rule provides a relatively simple method to approximate the post-action world by a subsuming affine-world. The problem of projecting a plan, i.e. a sequence of actions, on an affine-world can then be solved by sequentially applying PR3 to each action in the plan, yielding a final affine-world $w'$ that subsumes the actual final world $w$. The expected utility interval of the actual final world, $EUI(w)$ can then be bounded by $EUI(w')$, the EUI of the approximating affine-world, which can be efficiently computed using the recursive greedy knapsack algorithm (see the Affine-Worlds Utility Theorem). A plan can be eliminated if its expected utility interval is dominated by that of another plan.

The complexity of the process of evaluating and eliminating suboptimal plans depends mainly on three factors: (i) the number of the branches of the actions in a plan, (ii) the number of the actions in a plan, and (iii) the number of alternative plans in the plan space. Three abstraction techniques are discussed in [6] to reduce the complexity of the planning process. The *intra-action abstraction* technique, initially introduced by Hanks [12] who calls it *bundling*, reduces the branching factor of actions by replacing a set of branches by a single branch. We can use the Branch-Merging Lemma to apply this technique in our framework to produce correct abstractions. The *sequential-action abstraction* technique is used to reduce the length of a plan by replacing a subsequence of actions in that plan by a single action. We can use the Tree-Flatening Lemma to produce correct sequential abstractions. The *inter-action abstraction* technique, suggested by Tenenberg [14] is used to reduce the number of alternative plans in the plan space by grouping a set of actions (respectively plans) into a single action (respectively plan). We can use the Star-Merging Lemma to produce correct inter-action abstractions.

Before presenting the abstraction procedures within our framework, the notion of correct abstraction needs to be clarified. Action abstractions can be correct with respect to different projection rules. An abstract action $\Lambda^*$ is called a *correct inter-abstraction* of a set of actions $\{\Lambda_1, \Lambda_2, \ldots, \Lambda_n\}$ with respect to the projection rule number $j$, ($j = 1, 2, 3$) if $T_j(\Lambda^*, w) \supseteq T_j(\Lambda_i, w)$, for all $w \in \mathcal{AFF}$ and $i = 1, 2, \ldots, n$. The correctness criteria for intra- and sequential-action abstractions are defined similarly (see [7] for more details). Since plans cannot be projected with PR1, we can only define the correctness of abstract plans wrt to PR2 or PR3. An abstract plan $pl$ is a *correct inter-abstraction* of a set of plans $\{pl_1, pl_2, \ldots, pl_n\}$ wrt to projection rule number $j$ ($j = 2, 3$), if $T_j(pl, w) \supseteq pl_i(w)$, for all $w \in \mathcal{AFF}$ and $i = 1, 2, \ldots, n$.

It is not hard to see that an abstract plan that is built upon abstract actions that are correct wrt to a particular projection rule is correct wrt a looser projection rule. The problem here is to choose a projection rule to define the correctness criterion such that as-tight-as-possible abstraction procedures can be derived. As the result of a careful analysis, we conclude that the Second Projection Rule is the best for this purpose [9]. Due to space limitation, the full analysis of this problem is omitted here. Below we give the descriptions of the three abstraction procedures. The details are given in [7].

The three abstraction procedures are completely specified by three operators: *bundle-branches*, *combine-branches*, and *compose-branches*, respectively. Each of these operators takes as arguments two branches [10] $<C_1, P_1, E_1>$ and $<C_2, P_2, E_2>$ and produces a single branch $<C^*, P^*, E^*>$. The specification of $<C^*, P^*, E^*>$ for each operator is given below. We define the union of two effects $E_1$ and $E_2$ to be the function $E_1 \cup E_2(s) = E_1(s) \cup E_2(s)$, and the composition of $E_1$ and $E_2$ to be the function $(E_2 \circ E_1)(s) =$

---

[7] The convex hull of a set of real numbers is the smallest interval that contains all the numbers.

[8] The problem can be interpreted as the problem of packing objects of different values into a knapsack with fixed size (1) such that the total value is minimized (for the lower bound) or maximized (for the upper bound). This algorithm is from Doan [5, 4]. The greedy knapsack algorithm is also called the *annihilation/reinforcement* algorithm by Tessem [15], who calls the minimizing step *annihilation*, and the maximizing step *reinforcement*.

[9] Intuitively, the reason for this is the fact that PR1 produces non-convex sets, while PR3 is looser than PR2.

[10] It is straightforward to generalize these abstraction procedures to the case of grouping more than two branches (or actions).

$\cup_{e_2 \in E_2; e_1 \in E_1} e_2(e_1(s))$.

- *bundle-branches* (for the intra-action abstraction procedure): $C^* = C_1 \cup C_2$, $E^* = E_1 \cup E_2$, and $P^* = P_1 + P_2$ if $C_1 = C_2$ and $[\min\{L(P_1), L(P_2)\}, U(P_1) + U(P_2)]$ otherwise.

- *combine-branches* (for the inter-action abstraction procedure): $C^* = C_1 \cup C_2$, $E^* = E_1 \cup E_2$, and $P^* = [\min\{L(P_1), L(P_2)\}, \max\{U(P_1), U(P_2)\}]$ if $C_1 = C_2$ and $[0, \max\{U(P_1), U(P_2)\}]$ otherwise.

- *compose-branches* (for the sequential-action abstraction procedure): $C^* = C_1 \cap \{b \in \Omega \mid E_1(b) \cap C_2 \neq \emptyset\}$, $E^* = E_2 \circ E_1$, and $P^* = P_1 \times P_2$.

**Theorem 4** *The above action abstraction procedures are correct with respect to the Second Projection Rule.*

## 7    Related Work and Future Research

The affine-world representation in our framework has several advantages over the interval mass assignment representation of Doan [4, 5]. First, the affine-world representation is more general than the IMA representation; an IMA can be represented as a 2-level standard affine-tree, or a 1-level affine-star with leaves of the form $CH(B)$ (Theorem 1). During the process of projecting actions on an IMA, the post-action world must be approximated with an IMA obtained from flatening the projected affine-tree, incurring a loss of information which can be significant in some extreme cases. One may think that this loss of information is traded for some gain of simplicity (a 1-level tree, instead of a multi-level tree with the same number of leaves), but it is not the case for two reasons. First, maintaining a multi-level tree $T$ is not much more complex than a 1-level tree (star) $S$ with the same number of leaves. For example, if, on average, each node of $T$ has $k$ children, then, on average, the ratio of the total number of the nodes in $T$ to that in $S$ is: $(\sum_{i=0}^{r} k^i)/(1 + k^r) \approx k/(k-1)$, which approaches 1 if $k$ is large [11]. Second, since evaluating a plan often involves computing the utility of an *entire* chronicle [10], the planner in the IMA framework must associate each world in a chonicle with a separate IMA (1-level affine-tree), while the planner in our framework can comfortably encode the whole chronicle in a single multi-level tree.

Haddawy et al [9] have implemented the DRIPS planner, which uses the abstraction concepts discussed in this paper to construct and evaluate abstract plans. The DRIPS planner has been successfully applied to a number of real-world planning domains. The problems of estimating the loss due to abstraction and automatically generating abstraction hierarchies however remain quite difficult and challenging, and will form a main direction for our future work.

---

[11] Here we do not consider extreme cases such as the case when $T$ is a path. Flatening trees like this does not incur information loss.